\documentclass{article}

\usepackage{arxiv}

\usepackage[utf8]{inputenc} 
\usepackage[T1]{fontenc}    
\usepackage{hyperref}       
\usepackage{url}            
\usepackage{booktabs}       
\usepackage{amsfonts}       
\usepackage{nicefrac}       
\usepackage{microtype}      
\usepackage{lipsum}		
\usepackage{graphicx}
\usepackage{subcaption}
\usepackage{natbib}
\usepackage{doi}

\usepackage{ragged2e}
\usepackage{wrapfig}
\usepackage{color}

\title{Semantic match: Debugging feature attribution methods in XAI for healthcare}

\author{Giovanni Cin\`a$^{1,2, 3}$\\
	\texttt{g.cina@uva.nl} \\
	\And
	{Tabea E. R\"ober$^4$} \\
	\texttt{t.e.rober@uva.nl} \\
	\And
	{Rob Goedhart$^4$} \\
	\texttt{r.goedhart2@uva.nl} \\
 	\And
	{\c{S}. \.{I}lker Birbil$^4$} \\
	\texttt{s.i.birbil@uva.nl} \\
}

\date{}

\renewcommand{\headeright}{}
\renewcommand{\undertitle}{}
\renewcommand{\shorttitle}{Semantic match: Debugging feature attribution methods in XAI}

\begin{document}
\maketitle
\centering
$^1$ Department of Medical Informatics, Amsterdam University Medical Center, University of Amsterdam, NL

$^2$ Institute for Logic, Language, and Computation, University of Amsterdam, NL

$^3$ Pacmed, Amsterdam, NL

$^4$ Business Analytics, Amsterdam Business School, University of Amsterdam, Amsterdam, NL

\vspace{1cm}

\justifying

\begin{abstract}
The recent spike in certified Artificial Intelligence (AI) tools for healthcare has renewed the debate around adoption of this technology. One thread of such debate concerns Explainable AI (XAI) and its promise to render AI devices more transparent and trustworthy. A few voices active in the medical AI space have expressed concerns on the reliability of Explainable AI techniques and especially feature attribution methods, questioning their use and inclusion in guidelines and standards. Despite valid concerns, we argue that existing criticism on the viability of post-hoc local explainability methods throws away the baby with the bathwater by generalizing a problem that is specific to image data. We begin by characterizing the problem as a lack of semantic match between explanations and human understanding. To understand when feature importance can be used reliably, we introduce a distinction between feature importance of low- and high-level features. We argue that for data types where low-level features come endowed with a clear semantics, such as tabular data like Electronic Health Records (EHRs), semantic match can be obtained, and thus feature attribution methods can still be employed in a meaningful and useful way. Finally, we sketch a procedure to test whether semantic match has been achieved.
\end{abstract}


\section*{Introduction}
Along with the blooming of Artificial Intelligence (AI) and the accompanying increase in model complexity, there has been a surge of interest in explainable AI (XAI), namely AI that allows humans to understand its inner workings \cite[\textit{e.g.},][]{doshi-velez2017, linardatos2020, Gilpin.2018, biran2017, doran2017}. This interest is particularly keen in safety-critical domains such as healthcare, where it is perceived that XAI can engender trust, help monitoring bias, and facilitate AI development \cite[\textit{e.g.},][]{doshi-velez2017, lipton2018}. XAI has already shown to improve clinicians’ ability to diagnose and assess prognoses of diseases as well as assist with planning and resource allocation. For example, \citet{letham2015} developed a stroke prediction model that matches the performance of the most accurate Machine Learning (ML) algorithms, while remaining as interpretable as conventional scoring methods used in clinical practice.

There is a wide variety of techniques for XAI, and many categorizations have been proposed in the literature. Techniques can roughly be grouped into local vs. global, and model-specific vs. model-agnostic approaches. Local methods aim to explain model outputs for individual samples, while global methods focus on making models more explainable at an aggregate level. Model-specific methods are tailored to explain a specific type of model, while model-agnostic methods can be applied to a range of different models. Many of the well-known techniques yield post-hoc explanations, meaning that they generate explanations for already trained, so-called ‘black-box’, models. Alternatively, there exist approaches that are considered inherently explainable, also known as white-box (or glass-box) models, such as decision trees and linear regression models. A detailed taxonomy is beyond the scope of this paper; for an extensive overview we refer the reader to \citet{carvalho2019}, \citet{molnar_book2022}, and \citet{Ras.2022}.

A group of XAI techniques that has enjoyed substantial fame is the set of feature attribution methods, namely techniques that assign to each feature a measure of how much it contributes to the calculation of the outcome according to the model. Popular techniques produce such explanations in a local fashion, and among the most famous there are SHAP \cite{lundberg2017}, LIME \cite{Ribeiro.2016}, saliency maps \cite{selvaraju2017grad}, and integrated gradients \cite{sundararajan2017axiomatic}. To give a sense of the success of these techniques, it suffices to say that some of them are now integrated as default explainability tools in widespread cloud machine learning services, while in areas such as natural language processing, researchers are starting to use such methods as the gold standard against which they judge the quality of other explanations \cite{mohankumar2020towards}.

Despite the enthusiasm and a growing community of researchers devoting energy to XAI, there is currently no consensus on the reliability of XAI techniques, and several researchers have cast serious doubts on whether XAI solutions should be incorporated into guidelines and standards, or even deployed at all \cite[\textit{e.g.},][]{lipton2018, mccoy2021, Ghassemi.2021, neely2022song}.

\begin{wrapfigure}[35]{r}{0.4\textwidth}
  \centering
    \includegraphics[width=0.35\textwidth]{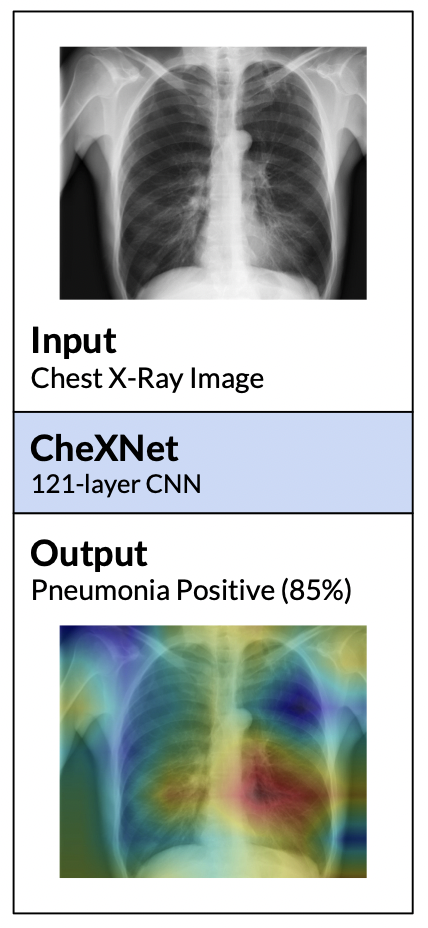}
  \caption{An example of visual explanation by heatmap for a medical image. Image courtesy of  [Rajpurkar et al., 2017].}
  \label{fig:fig1}
\end{wrapfigure}

Undoubtedly, there is an inherent tension between the desire for machines performing better than humans, and the requirement for machines to provide human-understandable explanations. Together with their super-human capacities –such as the ability to juggle dozens or hundreds of factors-- the sub-symbolic character of statistical learning techniques contributes to rendering many ML models opaque to humans. Simply put: humans cannot ‘read off’ what a neural network has learned just by looking at the matrix of weights.

The internal or ‘latent’ representation of a machine, namely the way in which the machine encoded the patterns found in the data, is what we would like to explain to the human, together with the way in which this internal representation interacts with a single data point to generate an output. The solution presented by feature attribution methods is also sub-symbolic, in the sense that an explanation also takes the form of a vector or matrix of values, and here lies the source of the problem. As different scholars pointed out \cite{Ghassemi.2021, rudin_stop2019}, the assignment of meaning to such explanations can be tricky, sometimes lulling the humans into a false sense of understanding while the explanations are in fact flawed or misleading. This issue is particularly sensitive when such explainability techniques are used in high-stakes environments such as healthcare.


Does this mean that feature attribution methods are altogether unreliable? In this article we argue that existing criticism on the viability of post-hoc local explainability methods throws away the baby with the bathwater by generalizing a problem that is specific to unstructured data such as images. We characterize the issue with feature attribution methods as a lack of semantic match between explanations and human understanding. To understand when semantic match can be obtained reliably, we introduce a distinction between feature importance of low- and high-level features. We argue that in the case of data types for which low-level features come endowed with clear semantics, such as tabular data like EHRs, semantic match is enabled and thus feature attribution methods can still be employed in a meaningful and useful way. As for high-level features, we present a conceptual procedure to test whether semantic match is achieved, paving the way for future operationalization of this test.

\section*{The criticism on local feature attribution methods}

In this section we expand on the problem that feature attribution methods are confronted with. Applied to images, explanations generated by feature attribution methods present themselves as heat maps or colored overlays, indicating the contribution of specific pixels to the prediction of the model on the input at hand. Intuitively, highlighted regions comprise pixels which were considered ‘important’ by the model (see Figure \ref{fig:fig1}). \textit{Prima facie}, one might be led to believe that this allows humans to check that the model is paying attention to the right elements of the image, therefore increasing our trust in the specific prediction and in the model more generally. 

\begin{figure}
    \centering
    \includegraphics[width=0.9\textwidth]{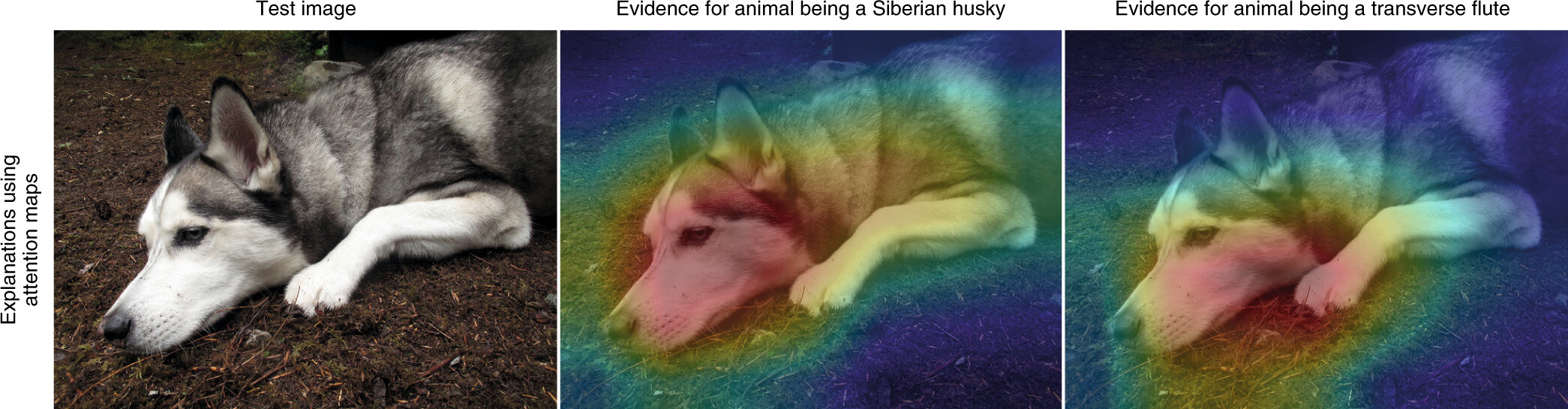}
    \caption{While the explanation of the first classification seems intuitive, this impression is put into question when a similar explanation is offered for an absurd classification. Figure reproduced from [Rudin 2019] with author's permission.}
    \label{fig:fig2}
\end{figure}

However, when a certain area of an image is highlighted we simply do not know if what we recognize, say the shape of a kidney or the beak of a bird, is the same as what the AI recognizes. As several researchers have pointed out, what look like plausible explanations at first may turn out to be ungrounded or spurious explanations when subjected to closer scrutiny. Figure \ref{fig:fig2} displays an example of this mishap: very similar explanations are offered for the prediction on two very different classes, invalidating our intuition that the model has learned to recognize dogs by their facial features.

This opens the door for different kinds of biases, both on the side of the user, who can project their beliefs erroneously onto the machine, and on the side of the XAI, which might provide feature attribution information that is inconsistent or misleading.

\begin{figure}
  \centering
  \begin{minipage}[t]{0.49\textwidth}
    \includegraphics[width=\textwidth]{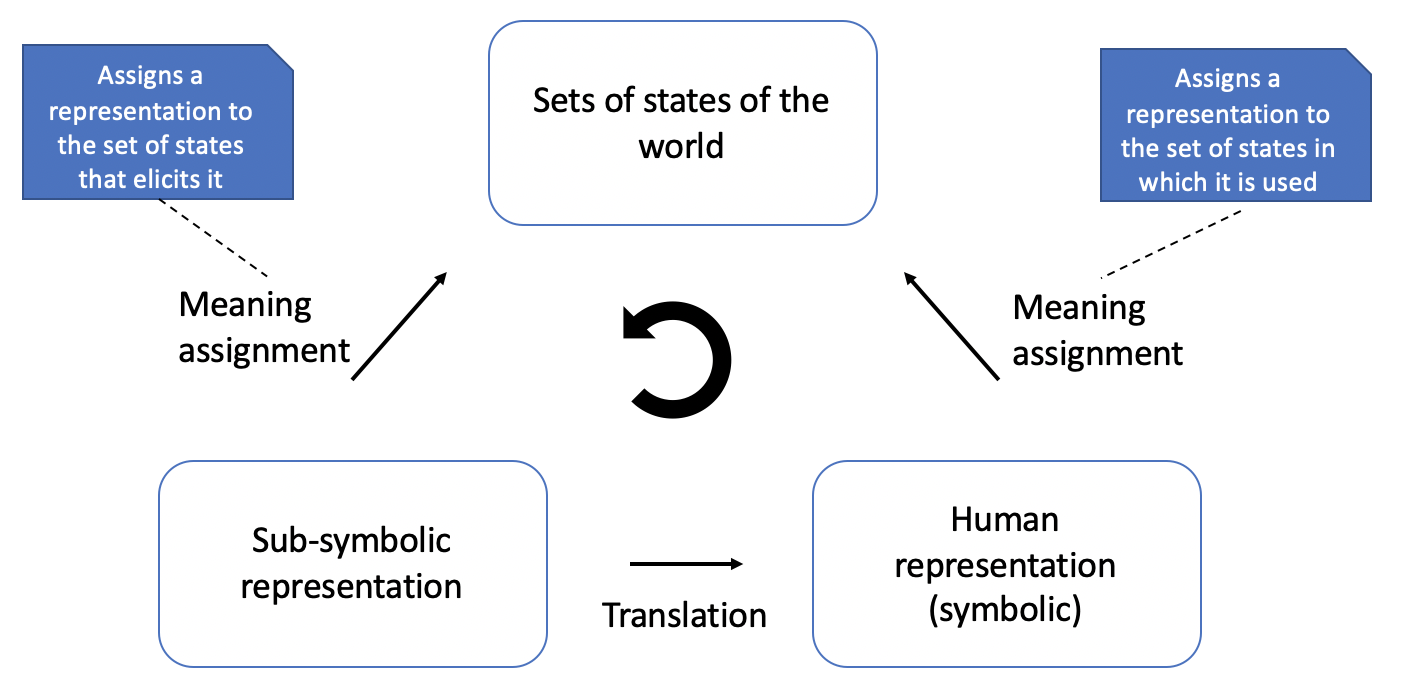}
    \caption{The diagram representing a semantic match.}
    \label{fig:fig3}
  \end{minipage}
  \begin{minipage}[t]{0.49\textwidth}
    \includegraphics[width=\textwidth]{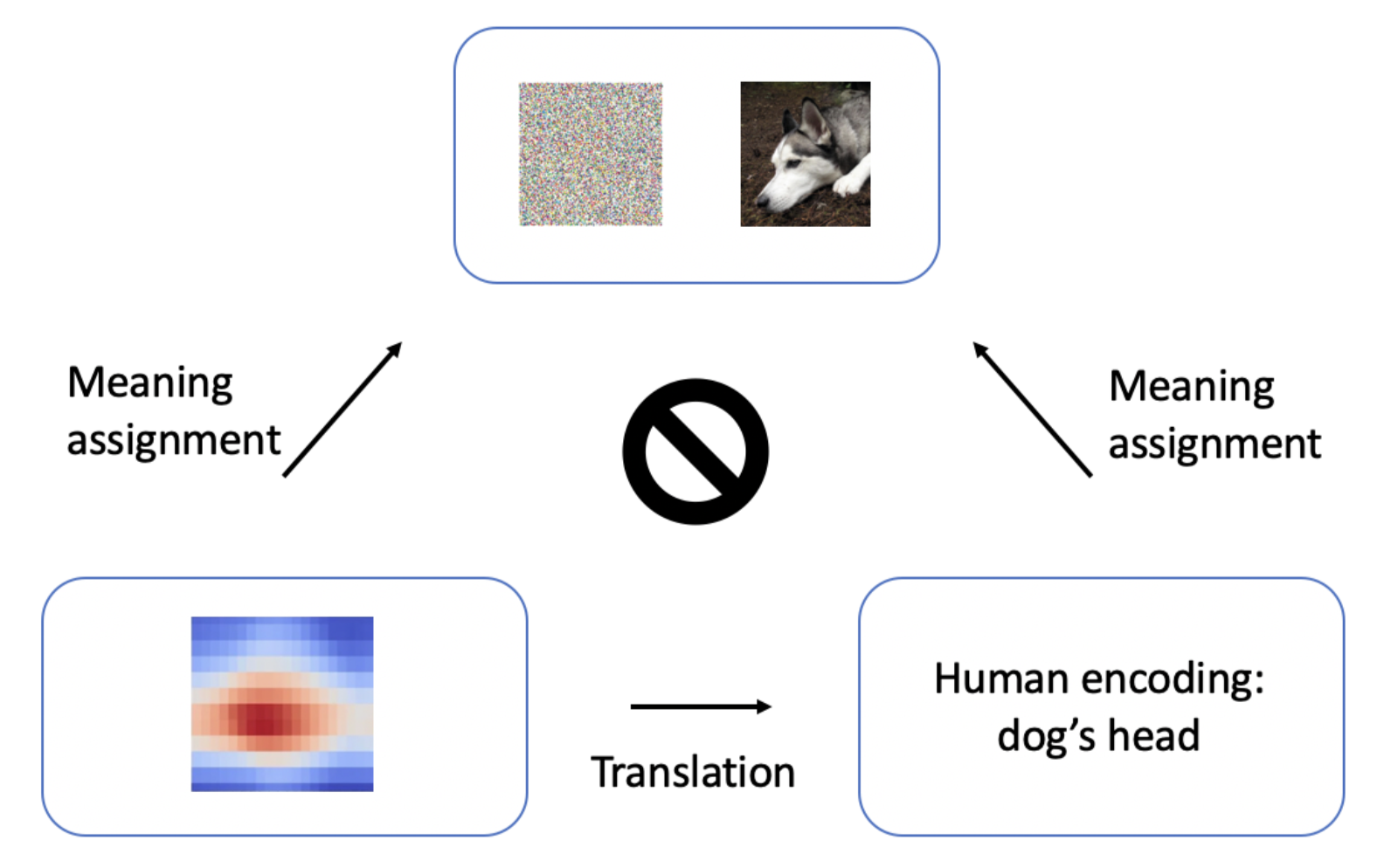}
    \caption{A scenario depicting a semantic mismatch. The images eliciting a certain explanation do not comply with our recollection of the explanation.}
    \label{fig:fig4}
  \end{minipage}
\end{figure}

The root of the problem, we maintain, lies in the inability of humans to attribute meaning to a sub-symbolic encoding of information. What is needed is a systematic way to translate sub-symbolic representations to human-understandable ones, in a way that respects how we assign meaning. This is represented schematically in Figure \ref{fig:fig3}. We call such a commuting diagram a \textbf{semantic match}. An ideal explanation would provide content of the bottom-left node and a suitable translation in order to obtain a semantic match: the explanation of a certain sub-symbolic representation encoded by the machine should have (i) a clearly defined meaning and (ii) an unambiguous way to translate to human terms with the same (or very similar) meaning. Semantic match, which is a crude simplification of complex cognitive and linguistic phenomena, offers a handy conceptual tool to debug explanations.

Indeed, just as we are unable to understand a latent representation, we are unable to relate to a heatmap unless it comes paired with a well-defined meaning assignment and translation, as illustrated in the diagram. Overlaying the heatmap to an image encourages us to use our visual intuition as translation, but alas this last step is an ill-advised one, since it gives us the illusion of a semantic match while in fact the explanations do not conform to expectations. Figure \ref{fig:fig4} exemplifies a failed semantic match. In this scenario a heatmap is generated as an explanation for the behavior of the model on an image. From the image, it may appear that the machine ‘recognizes’ a certain feature (a dog’s head), and therefore classifies the image as a specific class (‘dog’). However, another spurious input generates the same heatmap, and hence the translation is invalid and unreliable. To replicate the uncanny observation of Figure 2, one can construct another scenario where states of the world are pairs of input images and predictions. These are cases of semantic mismatch: the states of the world in which the heatmap is produced do not correspond to the states where one would plausibly use the concept of dog’s head to classify a dog image.

This criticism seems to undercut the utility of such local feature attribution methods: if they are potentially misleading and bring no clear added value, should they be used at all?


\section*{Distinguishing low- and high-level features}

In order to understand the utility of such methods, one needs to step back and consider what are the characteristics of the examples that give rise to the problem, and assess whether they generalize. Not unsurprisingly, most of the debate revolves around images, given the astounding success of deep learning in the realm of computer vision. Images are a prime example of what is sometimes called ‘unstructured data’, namely data that does not come equipped with additional structure and is akin to raw sensory data. Note that this is a misnomer, since such data typically is endowed with a rich structure (in the mathematical sense) deriving from the notion of distance defined between pixels or between parts of an ordered sequence. We will nonetheless adopt this terminology to facilitate readability, since it is ubiquitous.

One of the defining characteristics of images -and unstructured data more generally- is that a single feature has no intrinsic meaning: a certain color value of a pixel means nothing by itself. Only a pattern of values for a group of features can be attributed meaning, \textit{e.g.} a cloud of pixels with the shape and color of a dog’s head. In other words, we recognize properties or entities in an image by matching activation patterns with our visual intuitions. Following a widespread habit in the ML community, we will refer to patterns in groups of features as ‘high-level features’, while single features, i.e. the entries of the vector representing an input, will be dubbed ‘low-level features’ by contrast. With this terminology, we can succinctly state that in images only high-level features can be attributed meaning while low-level features cannot. 

On the contrary, in structured data each feature is usually conferred specific meaning; \textit{e.g.} in EHR data a certain value might contain the information pertaining to the blood pressure of a patient at a certain time. When such low-level features are not defined with a specific protocol - such as the ones for measurements of vital signs in clinical settings - they refer to standard concepts in natural language. We can, therefore, easily interpret what these low-level features mean regardless of the values of the other features. For instance, a value of 180 in the feature corresponding to systolic blood pressure gives us a piece of information that we can understand and process, even without knowing other features of a patient. 

To be sure, there are also high-level features in the case of structured data. Continuing with the example of EHRs, a high-level feature could be for example a phenotype which is not encoded explicitly as a feature but instead depends on the combination of existing features, say glucose, BMI, and so on. Hence, the crucial difference between structured and unstructured data is that the former has clear meaning for the low-level features, while on the high-level features the two data types behave similarly.

We maintain that the usage of a post-hoc feature attribution method hinges on the application of the corresponding semantic match diagram. Without a rigorously defined meaning and translation, a heatmap remains a matrix of values without rhyme or reason. We cannot extract information from such a matrix just as we cannot fathom what is encoded in the latent space of a neural network by eyeballing the value of a point in said space. So far the attribution of meaning to a heatmap-explanation has been carried out in an informal and intuitive way, which as previous scholars argued is prone to error and a false source of confidence.

\section*{Saving feature importance for low-level features}

The distinction between low-and high-level features helps untangle the cases in which semantic match works out-of-the box from those in which it fails. A post-hoc local feature attribution method can be used appropriately on low-level features when they have a predefined translation, as is the case for medical tabular data. For example, suppose a risk prediction model is trained on EHR data. When a patient is presented to the model, the model might provide a risk score associated with feature importance values for the patient’s lab values. Suppose systolic blood pressure is marked as the most important factor increasing the risk of a specific patient. In this case, there is no ambiguity: such level of importance is attributed to systolic blood pressure and nothing else, and any user with sufficient training knows what systolic blood pressure is. Note that this has nothing to do with how the importance is calculated (\textit{e.g.} if it takes into account feature interactions) or if it is a sensible level of importance; all we are stating is that the user can unambiguously understand what the importance is attributed to. In other words, the importance is attributed to something which semantically matches our concept of ‘systolic blood pressure’.

\begin{figure}

\centering
\begin{subfigure}[b]{0.75\textwidth}
   \includegraphics[width=1\linewidth]{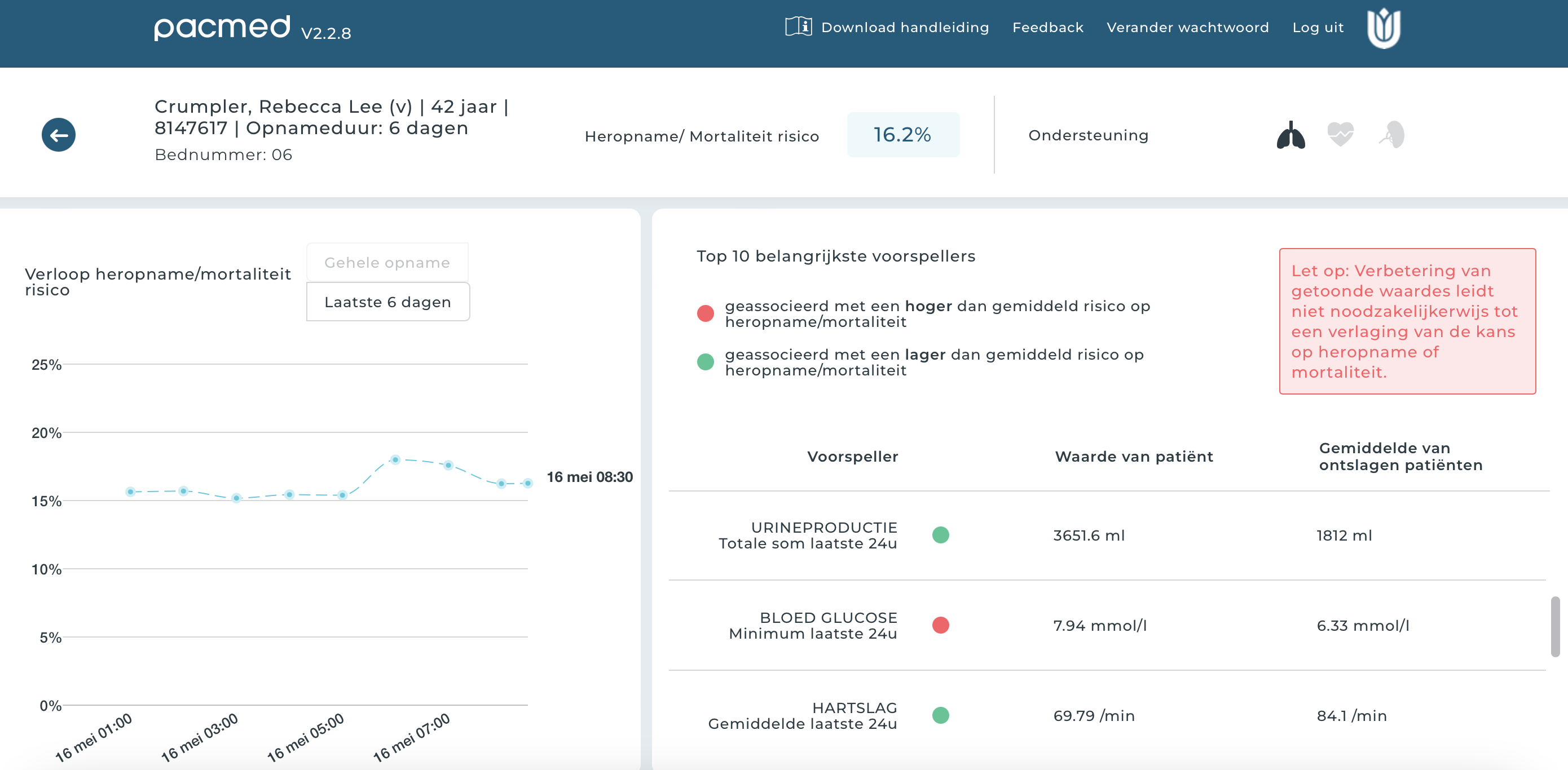}
   \caption{}
   \label{fig:Ng1} 
\end{subfigure}

\begin{subfigure}[b]{0.75\textwidth}
   \includegraphics[width=1\linewidth]{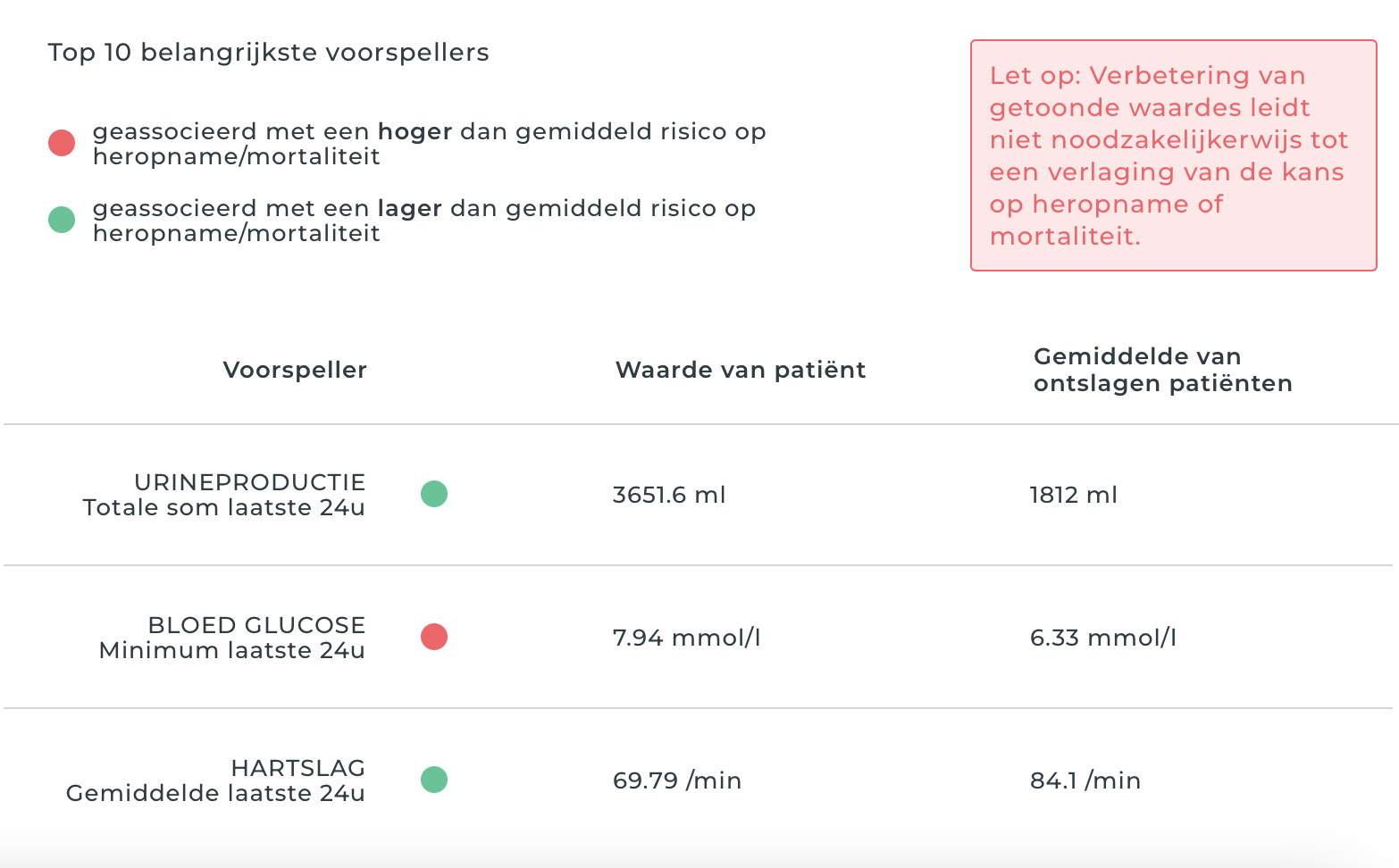}
   \caption{}
   \label{fig:Ng2}
\end{subfigure}
    
    \caption{(a) An example of the front-end of Pacmed Critical, a medical device currently used in the Amsterdam University Medical Center for the prediction of adverse outcomes after ICU discharge. (b) A close-up of the right side of the interface, displaying feature attribution for low-level features.  Courtesy of Pacmed; for background on the prediction model see \cite{thoral2021explainable, de2022predicting}. The displayed patient data is synthetic. }
    \label{fig:fig5}
\end{figure}

The user of such a model can then engage with the feature importance and assess whether it is sensible, while addressing questions like “Is this level of risk reasonable given a high importance of systolic blood pressure?” Such a question may be answered in the positive, if the clinician believes the value of systolic blood pressure is concerning, or in the negative, if for example, the clinician knows that the value of systolic blood pressure is a byproduct of medications she can control. An example of such an approach is displayed in Figure \ref{fig:fig5}, where the top features contributing to clinical risk are displayed along with a color code indicating if they are risk increasing or decreasing.

Crucially, semantic match allows the user of an explanation to engage with it, and to decide whether it is agreeable or not. In contrast, when the same risk prediction is trained on image data, this reasoning falls apart. What feature attribution can highlight are just high-level features, \textit{e.g.} the portion of an image with a kidney shape. But lacking a semantic match, clinicians cannot trust whether what they see in the image matches the machine’s internal representation. 

\section*{Debugging feature importance for high-level features}

Does this mean that we have to give up feature attribution for high-level features? In fact, what we need is a procedure to test whether semantic match is present or if it is violated. In what follows we sketch such procedure at a conceptual level.

Suppose a ML model $f$ has been trained on labeled data of the shape $(x, y)$, where $x$ represent an input vector and $y$ an output. We denote a local feature attribution method with $M$ and say that $M(f, x, y) = e$ is the explanation furnished by the model $f$ on such pair of input and output.
We are interested in testing whether we have semantic match with the explanation $e$, or in other words, if what we `see' in the explanation is indeed what the explanation is capturing. 

At an intuitive level, what we want to ascertain is that an explanation matches our translation of it. This is encoded in the commutation of the semantic diagram, namely that all the data points giving rise to a certain explanation are also complying to our translation hypothesis, which we will indicate with $\theta$, and vice versa. We are thus looking at answering two questions:
\begin{itemize}
    \item Is every input-output pair that generate an explanation similar to $e$ also a case in which $\theta$ is the  case?
    \item Is  every input-output pair in which $\theta$ is satisfied going to generate an explanation that is similar to $e$?
\end{itemize}

To exemplify the procedure, suppose one has developed an algorithm to classify pictures of animals. Presented with a picture classified as a dog and an explanation $e$, one formulates the translation hypothesis $\theta$ that the explanation highlights the tail of the animal. To answer the aforementioned questions, one would need to take images that generate explanations similar to $e$ and check whether the explanations of those samples highlight tails. For the second question, one would need to take images of animals with tails and consider how similar they are to $e$.

Note that in this procedure one may also select data points whose label is different from $y$. This may not be an issue, since the high-level feature used for the prediction of $x$ as $y$ may in theory also be used on another data point and another label (and hence be highlighted in the explanation). In the animal classification example, one may collect images of a panther in which a tail is displayed, and rule that the explanation correctly highlights a tail in those images.

What is interesting of this procedure is that it breaks down a theoretical issue in more concrete steps. A full formalization of the procedure and the demonstration of its implementation in practice are left for future work.

\section*{Discussion}

In this article we have reviewed the reliability problem of feature attribution methods and have proposed to diagnose the issue by means of the semantic match diagram. We have argued that without clear meaning and translation, semantic match cannot be obtained for high-level feature importance. A corollary of this statement is that current methods for feature attribution may not be appropriate for unstructured data unless semantic match is verified.

In contrast, structured data may still benefit from feature attribution, since for this data type low-level features have an in-built semantic match with human concepts. This allows humans to engage with explanations and exercise that all-important oversight that is required to spot failure modes of ML applications. Recent fairness concerns in the realms of human resources, healthcare, and law enforcement underscore the need for continued human control over semi-automated decisions.

When it comes to the limit of this analysis, it is important to remark that, even in the presence of semantic match, explanations can still fail to deliver on their promise. If an explanation is not faithful to the model, that is, it does not consistently represent the machine's behavior, the user may not leverage the explanation to agree or disagree with the machine’s output; recent work on this point shows a concerning divergence in explanations \cite{neely2022song}. 
It should also be added that not all data types neatly fall into the categories of structured and unstructured data. Textual data, for example, contains both tokens that have intrinsic meaning and tokens that have only contextual meaning, and therefore sits somewhat in the middle. Time series data is in a similar spot, exhibiting sequences whose single values may have defined meaning but whose evolution over time is harder to grasp. In these cases explanatory methods should be employed with caution and awareness of potential semantic mismatch. In the same vein, it should be recognized that, before the advent of neural networks that could process raw data, there was a long tradition of image processing by hand-crafting meaningful image features (see for example \cite{street1993nuclear}). In essence, such approaches built features with an intrinsic semantic match by encoding expert knowledge with feature engineering, turning unstructured data into structured one. These approaches may be rediscovered as ways to attribute meaning to explanations in consultation with domain experts. 
Finally, semantic match is a user-dependent concept: while users with the relevant background may correctly interpret an explanation, others may not, and in healthcare settings it is crucial to clearly identify user groups and provide proper training.

Beside operationalizing the procedure we introduced in the previous section, future work aiming to obviate problems deriving from failed semantic match could direct attention to generating explanations that comply with human categories by design, possibly even with special categories employed by the targeted user. One direction of future research explores the possibility to capture information on machine behavior in a symbolic manner by means of hybrid models \cite{Sarker.2021}. Another option might contemplate combining visual explanation with image segmentation, to fix the semantics of entities used in the explanation. Finally, more elaborate explanations may require access to ontologies regulating the relationship between entities (as in \textit{e.g.} \cite{Lecue.2018, liartis2021semantic}), which should themselves match ontologies used -- more or less implicitly -- by humans.

When it comes to medical AI, we want clinicians to be able to interact with machines in a meaningful way, namely with the right tools to adjudicate when the machine’s advice is worth following. Framing the problem in terms of semantic match helps shedding light on the issue that explanations are still too ambiguous and too far from clinicians’ reasoning. We should be building explanations in the clinician's language, rather than asking clinicians to rely on intuition or to learn to think like a computer scientist.

\bibliographystyle{unsrtnat}
\bibliography{main}  






\end{document}